\documentclass[english,runningheads,a4paper]{llncs}[2018/03/10]

\usepackage[ngerman,main=english]{babel}
\addto\extrasenglish{\languageshorthands{ngerman}\useshorthands{"}}
\usepackage{regexpatch}
\makeatletter
\edef\switcht@albion{%
  \relax\unexpanded\expandafter{\switcht@albion}%
}
\xpatchcmd*{\switcht@albion}{ \def}{\def}{}{}
\xpatchcmd{\switcht@albion}{\relax}{}{}{}
\edef\switcht@deutsch{%
  \relax\unexpanded\expandafter{\switcht@deutsch}%
}
\xpatchcmd*{\switcht@deutsch}{ \def}{\def}{}{}
\xpatchcmd{\switcht@deutsch}{\relax}{}{}{}
\edef\switcht@francais{%
  \relax\unexpanded\expandafter{\switcht@francais}%
}
\xpatchcmd*{\switcht@francais}{ \def}{\def}{}{}
\xpatchcmd{\switcht@francais}{\relax}{}{}{}
\makeatother

\usepackage{caption}
\usepackage{subcaption}
\usepackage{multicol}
\usepackage{multirow}
\usepackage{dsfont}
\usepackage{amsmath}
\usepackage{amssymb}
\usepackage{marvosym}

\usepackage[nottoc]{tocbibind}

\usepackage{footnote}
\makesavenoteenv{tabular}
\makesavenoteenv{table}

\usepackage{adjustbox}
\usepackage{textcomp}

\usepackage{booktabs} 
\usepackage{array}
\usepackage{arydshln}
\DeclareCaptionLabelFormat{andtable}{#1~#2  \&  \tablename~\thetable}

\usepackage{ifluatex}
\ifluatex
  \usepackage{fontspec}
  \usepackage[english]{selnolig}
\fi
\usepackage{newunicodechar}
\newunicodechar{°}{\degree}
\iftrue 
  \ifluatex
  \else
    \usepackage[%
      rm={oldstyle=false,proportional=true},%
      sf={oldstyle=false,proportional=true},%
      tt={oldstyle=false,proportional=true,variable=false},%
      qt=false%
    ]{cfr-lm}
  \fi
\else
  \ifluatex
  \else
    \usepackage{newtxtext}
    \usepackage{newtxmath}
    \usepackage[zerostyle=b,scaled=.9]{newtxtt}
  \fi
\fi

\ifluatex
\else
  \usepackage[T1]{fontenc}
  \usepackage[utf8]{inputenc} 
\fi

\usepackage{graphicx}

\usepackage{upquote}

\usepackage{booktabs}

\usepackage{paralist}

\usepackage{csquotes}

\usepackage{textcmds}

\usepackage{url}
\makeatletter
\g@addto@macro{\UrlBreaks}{\UrlOrds}
\makeatother

\usepackage{xcolor}

\usepackage{listings}
\renewcommand{\lstlistingname}{List.}
\usepackage{chngcntr}
\AtBeginDocument{\counterwithout{lstlisting}{section}}

\usepackage{pdfcomment}
\makeatletter
\newcommand{\printfnsymbol}[1]{%
  \textsuperscript{\@fnsymbol{#1}}%
}
\makeatother

\usepackage[%
  square,        
  comma,         
  numbers,       
  sort&compress, 
]{natbib}

\ifluatex
\else
\fi

\usepackage{stfloats}
\fnbelowfloat

\usepackage{hyperref}
\hypersetup{hidelinks,
  colorlinks=true,
  allcolors=black,
  pdfstartview=Fit,
  breaklinks=true}
\usepackage[all]{hypcap}

\usepackage[group-four-digits,per-mode=fraction]{siunitx}

\usepackage[capitalise,nameinlink]{cleveref}
\usepackage{iflang}
\IfLanguageName{ngerman}{
  \crefname{table}{Tab.}{Tab.}
  \Crefname{table}{Tabelle}{Tabellen}
  \crefname{figure}{\figurename}{\figurename}
  \Crefname{figure}{Abbildungen}{Abbildungen}
  \crefname{equation}{Gleichung}{Gleichungen}
  \Crefname{equation}{Gleichung}{Gleichungen}
  \crefname{listing}{\lstlistingname}{\lstlistingname}
  \Crefname{listing}{Listing}{Listings}
  \crefname{section}{Abschnitt}{Abschnitte}
  \Crefname{section}{Abschnitt}{Abschnitte}
  \crefname{paragraph}{Abschnitt}{Abschnitte}
  \Crefname{paragraph}{Abschnitt}{Abschnitte}
  \crefname{subparagraph}{Abschnitt}{Abschnitte}
  \Crefname{subparagraph}{Abschnitt}{Abschnitte}
}{
  \crefname{section}{Sect.}{Sect.}
  \Crefname{section}{Section}{Sections}
  \Crefname{figure}{Fig.}{Figures}
  \crefname{figure}{figure}{figures}
  \crefname{listing}{\lstlistingname}{\lstlistingname}
  \Crefname{listing}{Listing}{Listings}
}

\usepackage{xspace}

\DeclareFontFamily{U}{MnSymbolC}{}
\DeclareSymbolFont{MnSyC}{U}{MnSymbolC}{m}{n}
\DeclareFontShape{U}{MnSymbolC}{m}{n}{
  <-6>    MnSymbolC5
  <6-7>   MnSymbolC6
  <7-8>   MnSymbolC7
  <8-9>   MnSymbolC8
  <9-10>  MnSymbolC9
  <10-12> MnSymbolC10
  <12->   MnSymbolC12%
}{}
\DeclareMathSymbol{\powerset}{\mathord}{MnSyC}{180}

\ifluatex
\else
  \input glyphtounicode
\fi

\begin{document}

\title{AutoSNAP: Automatically Learning Neural Architectures for Instrument Pose Estimation}
\titlerunning{AutoSNAP: Auto. Learning Neural Architectures for Instr. Pose Estim.}

\author{David Kügler\thanks{equal contribution}\inst{,1,2}\and 
	Marc Uecker\printfnsymbol{1}\inst{,1}\and
	Arjan Kuijper\inst{1,3} \and
	Anirban Mukhopadhyay\inst{1}}
\authorrunning{David Kügler, Marc Uecker, Arjan Kuijper, Anirban Mukhopadhyay}
\institute{
	Department of Computer Science, TU Darmstadt, Darmstadt, Germany\and
	German Center for Neuro-degenerative Diseases (DZNE), Bonn, Germany\and
	Fraunhofer IGD, Darmstadt, Germany\\
	\email{david.kuegler@dzne.de}
}

%
%
%
%
%
\maketitle
\begin{abstract}
Despite recent successes, the advances in Deep Learning have not yet been fully translated to Computer Assisted Intervention (CAI) problems such as pose estimation of surgical instruments. 
Currently, neural architectures for classification and segmentation tasks are adopted ignoring significant discrepancies between CAI and these tasks. 
We propose an automatic framework (AutoSNAP) for instrument pose estimation problems, which discovers and learns 
architectures for neural networks.  
We introduce
1)~an efficient testing environment for pose estimation, 
2)~a powerful architecture representation based on novel Symbolic Neural Architecture Patterns (SNAPs), and 
3)~an optimization of the architecture using an efficient search scheme.
Using AutoSNAP, we discover an improved architecture (SNAPNet) which outperforms both the hand-engineered i3PosNet and the state-of-the-art architecture search method DARTS.
\end{abstract}
\section{Introduction}\label{sec:intro}
Deep Neural Networks (DNNs) have revolutionized Computer-Assisted Interventions (CAI) with applications ranging from instrument tracking to quality control \cite{MaierHein.2017,Vercauteren.2020}.
However, the design of these neural architectures is a time-consuming and complex optimization task requiring extensive hyper-parameter testing. 
Consequently, CAI researchers often adopt established neural architectures designed for other vision tasks such as large-scale image classification \cite{Twinanda.2017,Hajj.2018,Miao.2016, Unberath.2019}.
But CAI problems requiring regression instead of classification or segmentation on scarcely annotated and small datasets differ from these tasks on a fundamental level.
This CAI-centric challenge is featured in instrument pose estimation for minimally-invasive temporal bone surgery \cite{Schipper.2004}: 
Training on synthetic data is necessary, because hard-to-acquire real-world datasets with high-quality annotation are reserved for evaluation only.
The state-of-the-art method (i3PosNet \cite{Kugler.2020}) relies on an architecture optimized for classification. 
It disregards the specialization potential as described by the no-free-lunch-theorem and as realized by DNNs for registration \cite{Balakrishnan.2019} demanding a method to automatically improve the architecture.

Optimizing neural architectures for a specific problem is challenging on its own due to the following requirements: 
1)~an \textit{Efficient Environment} to test candidate performance, 
2)~a \textit{Succinct Representation} to describe the architecture, and 
3)~an \textit{Efficient Search Algorithm} to improve candidates quickly with limited hardware.
Automatic Neural Architecture Search (NAS) strategies were initially introduced in computer vision classification.
Previous work \cite{Elsken.2019} can be classified into two groups: discrete and continuous.
The discrete strategy (e.g. NASNet) \cite{Zoph.2017,Luo.2018} iteratively proposes, tests and improves blocks. 
These blocks consist of multiple “NAS units” and are themselves combined to form full architectures. 
Despite being widely used in various publications, these units are not particularly efficient in terms of both optimization and functional redundancy. 
However, the iterative improvement scales well for distributed computing with massive computational effort (>=200 GPU days).
The continuous strategy (e.g. DARTS \cite{Liu.2019}) stacks all layer options together and calculates a weighted sum, motivating the name continuous. 
All architectures are trained at the same time and weights are shared. This approach is more computationally efficient (4 GPU days), but very VRAM-demanding because of “stacks of layers”.
AutoSNAP combines the flexibility of NASNet with the speed of DARTS by introducing an intuitive yet succinct representation (instead of NAS units) and improving the efficient search and optimization strategy.
The medical imaging community has recently confirmed the potential of NAS methods to segmentation \cite{Dong.2019b,Zhu.2019,Yu.20191220,Weng.2019} with adaptations for scalable \cite{Kim.2019} and resource-constrained \cite{Bae.2019} environments.
We are not aware of any application of NAS to CAI.
%

We introduce problem-dependent learning and optimization of neural architectures to instrument pose estimation. \footnote{We will publish our code at \url{https://github.com/MECLabTUDA/AutoSNAP}.}.
AutoSNAP implements problem-specific and limited-resources NAS for CAI with three major contributions:
1)~the integration of a CAI-framework as an \emph{efficient testing environment} for performance analysis (\Cref{fig:overview:a}),
2)~an extensible, \emph{succinct representation} termed Symbolic Neural Architecture Pattern (SNAP, \Cref{fig:overview:b}) to describe architecture blocks, and
3)~an \emph{efficient search algorithm} guided in ``Optimization Space'' (auto-encoder latent space) to explore and discover ``new architectures'' (\Cref{fig:overview:c}).
By integrating these factors, AutoSNAP links architecture and performance allowing for end-to-end optimization and search.
We jointly train AutoSNAP's auto-encoder (\Cref{fig:overview:c}) using a multi-component loss.
In addition to reconstruction, this loss also uses on-the-fly performance metrics from the testing environment to predict the performance of a SNAP-based architecture.
In consequence, we enable the substitution of the optimization on SNAPs by the optimization in a traversable ``Optimization Space''.
We show experimentally, that our automated approach produces improved architecture designs significantly outperforming the non-specialized state-of-the-art design. 
Additionally, AutoSNAP outperforms our reimplementation of the state-of-the-art NAS method DARTS \cite{Liu.2019} for pose estimation of surgical instruments (DARTS$^*$).
\section{Methods}
\label{sec:methods}
{
\setlength{\belowcaptionskip}{0pt}%
\setlength{\abovecaptionskip}{0pt}%
\setlength{\textfloatsep}{6pt}%
\setlength{\intextsep}{0pt}%
\begin{figure}[t!]
	\centering
	\begin{subfigure}{\linewidth}
		\centering
		\includegraphics[trim=0 -30 0 0,clip,width=\textwidth]{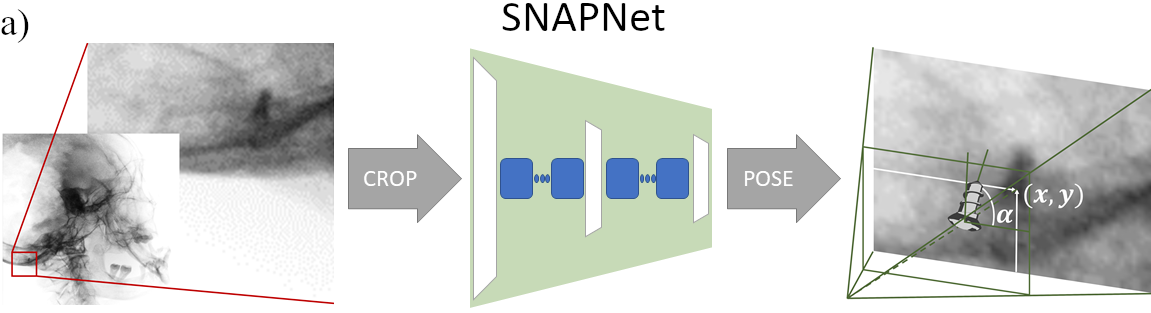}
		\caption{\emph{Testing Environment} from CAI: we search for a performant architecture (SNAPNet, green architecture) to estimate the pose of a surgical instrument from X-ray images.\vspace{10pt}}\label{fig:overview:a}
	\end{subfigure}

	\begin{subfigure}{\linewidth}
		\centering
		\includegraphics[trim=0 0 0 0,clip,width=\textwidth]{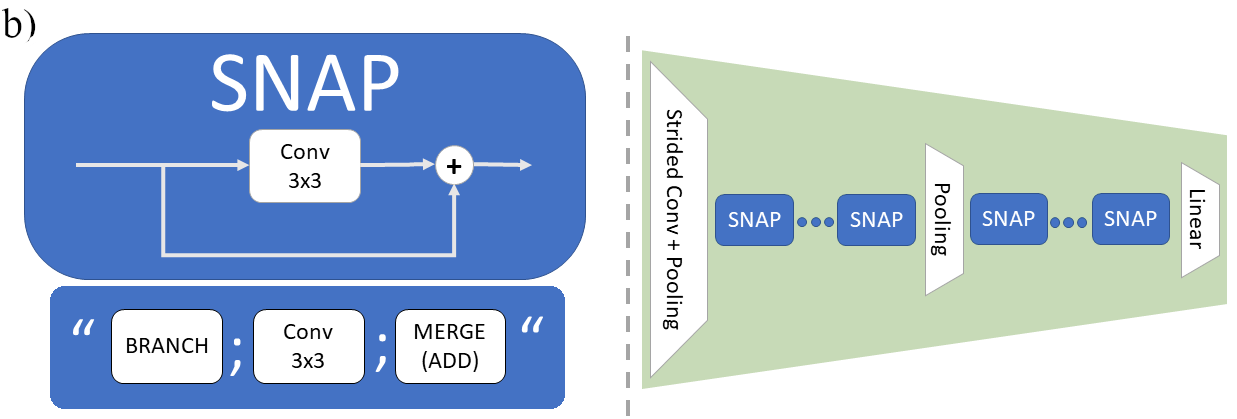}
		\caption{\emph{Succinct Representation}: left side: a SNAP (top) defines a corresponding neural block (bottom); right: we build the architecture (SNAPNet) by repeating this block.\vspace{10pt}}\label{fig:overview:b}
	\end{subfigure}
	\begin{subfigure}{\linewidth}
		\centering
		\includegraphics[trim=0 0 0 0,clip,width=0.8\textwidth]{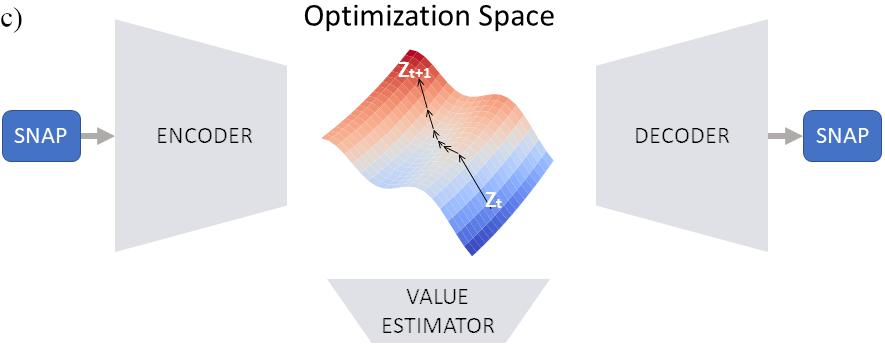}
		\caption{\emph{Efficient Search}: the transformation of SNAPs into a unified latent-space (auto-encoder) accelerates the search by gradient ascent on the value estimator surface.}\label{fig:overview:c}
	\end{subfigure}
	\caption{Overview of AutoSNAP components}
	\label{fig:overview}
\end{figure}%
}%
Here, we present the details of pose estimation (environment), SNAP (succinct representation), the auto-encoder and the optimization scheme (efficient search).
\subsection{Problem definition of pose estimation}
\label{ssec:problemdefinition}
\begin{figure}[!t]
	\centering
	\parbox{58mm}{%
		\includegraphics[trim=0 0 0 0, clip,width=56mm]{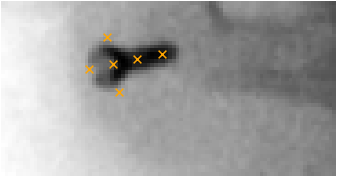}
	}\begin{minipage}{63mm}%
		\centering%
		\begin{tabular}{p{28mm}p{28mm}}
			\textbf{layer} & \textbf{topology}\\
			\textbf{symbols} & \textbf{symbols}\\
			\midrule
			\texttt{Conv 1x1} & \texttt{branch}\\
			\texttt{Conv 3x3} & \texttt{merge\,(add)} \\
			\texttt{DW-Conv 3x3} & \texttt{switch}\\
			\texttt{DWS-Conv 3x3} \\
			\multicolumn{2}{l}{\texttt{Max-Pool 3x3} (stride 1)}\\
			\bottomrule
		\end{tabular}
	\end{minipage}
	
	\parbox{53mm}{%
		\caption{\label{fig:neuralproblem}X-ray image patch of a screw with ``virtual landmarks''.}
	}\hspace{5mm}~\begin{minipage}{60mm}%
		\captionof{table}{\label{tab:symbols}SNAP symbols: \texttt{Conv}: Convolution, \texttt{DW}: Depthwise, \texttt{DWS}: DW-Separable}
	\end{minipage}
\end{figure}

To guide surgeons and robotic instruments in image-guided temporal bone surgery, instrument poses need to be estimated with high-precision.
Since the direct prediction of poses from full images is difficult, the state-of-the-art modular framework i3PosNet \cite{Kugler.2020} implements ``CROP'' and ``POSE'' operations (see \Cref{fig:overview:a}).
These simplifications significantly stabilize the learning problem by converting it to a patch-based prediction of ``virtual landmarks''.
``CROP'' uses a rough initial pose estimate to extract the region of interest, 
``POSE'' geometrically reconstructs the 3D pose of surgical instruments from six ``virtual landmarks''.
\Cref{fig:neuralproblem} shows a patch from a real X-ray image and predicted landmarks. 
i3PosNet then iterates these operations using earlier prediction as improved estimates.

Framing this prediction task as our \emph{environment}, we search for a neural architecture (green network in \Cref{fig:overview}) that minimizes the Mean-Squared Error of the point regression task (regMSE).
i3PosNet, on the other hand, only adapts a non-specialized VGG-based architecture for this task.
Our implementation parallelizes training and evaluation on a validation dataset on multiple machines.

\subsection{Symbolic Neural Architecture Patterns (SNAPs)}
\label{ssec:snap}
With many layer types and the design of connections in DNNs, it is currently impractical to optimize the topology and layer choice of the full architecture.
As a result, our full architecture (SNAPNet) repeats a block representing multiple operations as illustrated in \Cref{fig:overview:b}.
The topology and layer choice of this block are defined by a SNAP sequence.

To automatically generate trainable models, we introduce a language to define blocks using 8 SNAP symbols (see \Cref{tab:symbols}).
Each symbol corresponds to a modification of a stack of tensors which is used to build the model.
Five symbols specify (trainable) layers (\Cref{tab:symbols}).
Results replace the top tensor using the previous value as input.
Convolutions are always preceeded with BatchNormalization and ReLU activation.
The three topology symbols realize modification of the stack size and order for example enabling skip connections (see \Cref{fig:overview:b}, left).
\texttt{branch} duplicates the top element on the stack, \texttt{switch} swaps the top two elements and \texttt{merge\,(add)} pops the top two inputs, applies \texttt{concat} + \texttt{Conv 1x1} and pushes the result.
The stack is initialized by the output of the last two blocks and 
SNAPs end with an implicit \texttt{merge\,(add)} across all tensors on the stack (ignored in \Cref{fig:overview:b} for simplicity, but \Cref{fig:SNAP} includes these).

\subsection{AutoSNAP's auto-encoder}
\label{ssec:controller}
We introduce an auto-encoder architecture to transform the SNAP sequence into a 16-dimensional vector (latent space), since continuous vectors have favorable properties for optimization.
In addition to the Encoder and Decoder, a Value Estimator predicts the architecture performance (\Cref{fig:overview:c}).

The \textbf{Encoder} ($E$) and the \textbf{Decoder} ($D$) use a Recurrent Neural Networks with mirrored architectures of two bi-directional LSTMs and two fully connected layers\footnote{We provide additional diagrams of the architectures in the Supplementary Materials.}.
Since the last Encoder-layer uses tanh-activation, the latent space is bound to the interval of [-1,1] in each dimension.
As a conditional language model, the decoder generates a sequence of symbol probabilities from a latent vector.
Finally, the \textbf{Value Estimator} ($V$) is a linear regression layer with no activation function.
Since we are interested in both high accuracy and resolution for well-performing architectures (i.e. at very low regMSE values),
we estimate $-\log_{10}(\text{regMSE})$ of the candidate architecture on the validation dataset.
This value metric improves the resolution and gradients of the value estimator.

We train the auto-encoder on three sets of input and target data: 
1)~Sequence Reconstruction, 2)~Latent-space consistency and 3)~Value Regression.
1)~A Cross-Entropy (CE) loss enforces successful \emph{reconstruction} on randomly generated SNAP sequences ($\hat{X}$).
2)~The \emph{consistency of the latent space} is further supported by mapping uniform random latent vectors ($\hat{z}$) to a sequence of symbol probabilities, back to the latent space under a Mean Squared Error (MSE) loss.
3)~Value estimator and encoder are jointly trained to estimate the value criterion ($Y$) of known SNAPs ($X$) via MSE loss.
All three loss functions are minimized simultaneously via Gradient Descent:
\begin{equation} 
\mathcal{L} = \mathcal{L}_{\mathrm{CE}}(D(E(\hat{X})),\hat{X})+\mathcal{L}_{\mathrm{MSE}}(V(E(X)),Y)+\mathcal{L}_{\mathrm{MSE}}(E(D(\hat{z})),\hat{z})
\label{eqn:loss}
\end{equation}
Three failure modes motivate this design:
1)~A disentanglement between value estimate and architecture description occurs, rendering optimization within the latent space futile.
2)~The encoder only projects to a limited region of the latent space, thus some latent vectors have an unconstrained value estimate and no corresponding architecture (see~a).
3)~The decoder overfits to known architectures, which leaves a strong dependence on initial samples.
The Latent-space consistency enforces a bijective mapping between latent and symbol spaces (cycle consistency loss) addressing issues 1 and 2.
We mitigate problem 3 by only training the auto-encoder on randomly generated sequences.

\subsection{Exploration \& Optimization}
\label{ssec:optimization}
Our approach introduces the continuous optimization of an architecture block by gradient ascent inside the latent space of the auto-encoder (\Cref{fig:overview:c}).
The gradient for this optimization is provided by the value estimator which predicts the performance of the architecture pattern vector of a corresponding block (i.e. SNAP).
By the design of the auto-encoder, the latent space is learned to represent a performance-informed vector representation.
The thereby improved convexity of the latent space w.r.t. performance is the central intuition of the search.

The architecture search\label{chap:methods:opt}
consists of three iterative steps:
1)~Retrain the auto-encoder on all previously evaluated SNAPs (initially this is  a random sample of evaluated architectures).
2)~Transform a batch of best known SNAPs into the latent space, 
optimize them via gradient ascent on value estimator gradients until the decoded latent vector no longer maps to a SNAP with known performance.
After a limit of 50 gradient steps (i.e. in vicinity of the SNAP all architectures are evaluated), a new latent vector is sampled randomly (uniform distribution) from the latent space and gradient ascent resumed.
3)~Evaluate SNAPs found in step 2) using the test environment and add them to the set of known architectures.
Finally, repeat the iteration at step 1).
\section{Experiments}
\label{sec:experiments}
We evaluate our method on synthetic and real radiographs of surgical screws in the context of temporal bone surgery \cite{Schipper.2004}.
In this section, we summarize data generation, the parameters of the AutoSNAP optimization and settings specific to pose estimation as well as the evaluation metrics.



\subsection{Training and Evaluation Datasets}
\label{ssec:dataset}
We use the publicly available i3PosNet Dataset \cite{Kugler.2020} assuming its naming conventions. Dataset A features synthetic and Dataset C real radiographs.
We train networks exclusively on Dataset A, while evaluating on synthetic and real images.
\ \\
\textbf{Dataset A: Synthetic Images}: 
The dataset consists of 10,000 digital images (Subjects 1 and 2) for training and 1,000 unseen images (Subject 3) for evaluation with geometrically calculated annotations.
Training images are statically augmented 20-fold by random shifts and rotations to ensure similarity between training runs.
For architecture search, we split the training dataset by 70/10/20 (training/online validation/testing and model selection) to identify the performance of candidate models without over-fitting to the evaluation dataset.
\ \\
\textbf{Dataset C: Real Images}: Real X-ray images of medical screws on a phantom head are captured with a Ziehm c-arm machine (totaling 540 images).
Poses are manually annotated with a custom tool.

\subsection{Details of Optimization}
\label{ssec:controllersettings}
We randomly choose 100 SNAP architectures for the initial training of the autoencoder and the value-estimator. 
During the search phase, 100 additional models are tested by training the models for 20 epochs and evaluation on the validation set.
We stop the search after 1500 tested models.
Our small-scale test environment uses 4 blocks with a pooling layer in the center.
Convolutions have 24/48 features before/after the pooling layer.
In total, the search takes 100 GPU hours (efficiently parallelized on two NVIDIA GeForce GTX 1080 Ti for \SI{50}{\hour}) and requires no human interaction.

\subsection{DARTS architecture search}
\label{ssec:darts}
We compare our results with DARTS${}^*$, our reimplementation of DARTS \cite{Liu.2019} where the $^*$ indicates our application to CAI.
DARTS is an efficient, state-of-the-art NAS approach for classification from computer vision (CIFAR-10). 
We ported the DARTS framework to tensorflow implementing all operations as documented by the authors.
This process yields a large ``continuous model'' with weights for the contribution of individual layers.
For the evaluation and comparison with SNAPNet, we discretize and retrain the continuous model of DARTS$^*$ in analogy to the DARTS transfer from CIFAR-10 to ImageNet.
Similar to DARTS on ImageNet, our DARTS${}^*$ search took approximately 4 days on one GPU. 
Inherently, DARTS cannot efficiently be parallelized across multiple GPUs or machines because all updates are applied to the same continuous model.

\begin{table*}[t]
\begin{adjustbox}{center}
\begin{tabular}{ll|rr||rr}\toprule
	& & \multicolumn{2}{c||}{Dataset A: synthetic images} & \multicolumn{2}{c}{Dataset C: real images} \\
&\textbf{Model} & \textbf{Position [mm]} & \textbf{Angle\,[deg.]} &\textbf{Position [mm]} & \textbf{Angle [deg.]} 
\\ \midrule
\parbox[t]{4mm}{\multirow{4}{*}{\rotatebox[origin=c]{90}{3 iterations}}}
&i3PosNet \cite{Kugler.2020} & $0.024 \pm 0.016$ & $0.92 \pm 1.22$ & $1.072 \pm 1.481$ & $9.37 \pm 16.54$
\\
&DARTS${}^*$ \cite{Liu.2019} & $0.046 \pm 0.105$ & $1.84 \pm 6.00$ & $1.138 \pm 1.199$ & $9.76 \pm 18.60$
\\  \cline{2-6}
&SNAPNet-A (Ours)& $0.017 \pm 0.012$ & $0.52 \pm 0.88$ & $0.670 \pm 1.047$ & $7.55 \pm 14.22$
\\
&\textbf{SNAPNet-B (Ours)} & $\mathbf{0.016} \pm 0.011$ & $\mathbf{0.49} \pm 0.84$ & $\mathbf{0.461} \pm 0.669$ & $\mathbf{5.02} \pm 9.28$
\\ \midrule[1pt]
\parbox[t]{2mm}{\multirow{4}{*}{\rotatebox[origin=c]{90}{1 iteration}}}
&i3PosNet \cite{Kugler.2020} & $0.050 \pm 0.139$ & $1.14 \pm 1.50$ & $0.746 \pm 0.818$ & $6.59 \pm 10.36$
\\
&DARTS${}^*$ \cite{Liu.2019} & $0.062 \pm 0.146$ & $1.81 \pm 4.20$ & $0.810 \pm 0.770$ & $7.68 \pm 12.70$
\\  \cline{2-6}
&SNAPNet-A (Ours) & $0.026 \pm 0.029$ & $0.72 \pm 1.19$ & $0.517 \pm 0.678$ & $5.32 \pm 8.85$
\\
&\textbf{SNAPNet-B (Ours)} & $\mathbf{0.025} \pm 0.028$ & $\mathbf{0.65} \pm 1.06$ & $\mathbf{0.419} \pm 0.486$ & $\mathbf{4.36} \pm 6.88$
\\\bottomrule
\end{tabular}
\end{adjustbox}
\caption{Pose Evaluation (lower is better) for Datasets A and C (synthetic and real X-ray images). Evaluation for one and three iterations of the i3PosNet-scheme. Mean Absolute Error $\pm$ Standard Deviation of the absolute error.}
\label{tab:results}
\end{table*}
\subsection{Full-scale Retraining}
We retrain the full final architecture on the common training data of synthetic images from Dataset A \cite{Kugler.2020}.
Since efficiency is not a constraining factor for full training, we increase the number of blocks to a total of eight, four before and after the central pooling layer (see \cref{fig:overview:b}).
While SNAPNet-A uses 24/48 feature channels (same as the test environment), we increase the number of features to 56/112 (before/after the pooling layer) for SNAPNet-B.
In consequence, the number of weights approximately quadruple from SNAPNet-A and the discrete DARTS$^*$ model to SNAPNet-B and again to i3PosNet.
Like i3PosNet, we train models for 80 epochs, however using RMSProp instead of Adam.
Following the spirit of automatic machine learning, we obtained hyperparameters for these models using bayesian optimization.

\subsection{Evaluation Metrics}
To maximize comparability and reproducibility, we follow the evaluation protocol introduced by i3PosNet \cite{Kugler.2020}.
Similar to i3PosNet, we report mean and standard deviation of the absolute error for position and forward angle.
These are calculated w.r.t. the projection direction ignoring depth.
The forward angle is the angle between the image's x-axis and the screw axis projected into the image (\Cref{fig:overview:a}).
The architecture optimization performance and effectiveness is reported by the value metric ($-\log_{10}(\text{regMSE})$, see \Cref{ssec:controllersettings}).

\section{Results}
\label{sec:results}
\begin{figure}[t]
	\centering
	\parbox{68mm}{%
		\includegraphics[trim=28 5 50 30, clip,width=68mm]{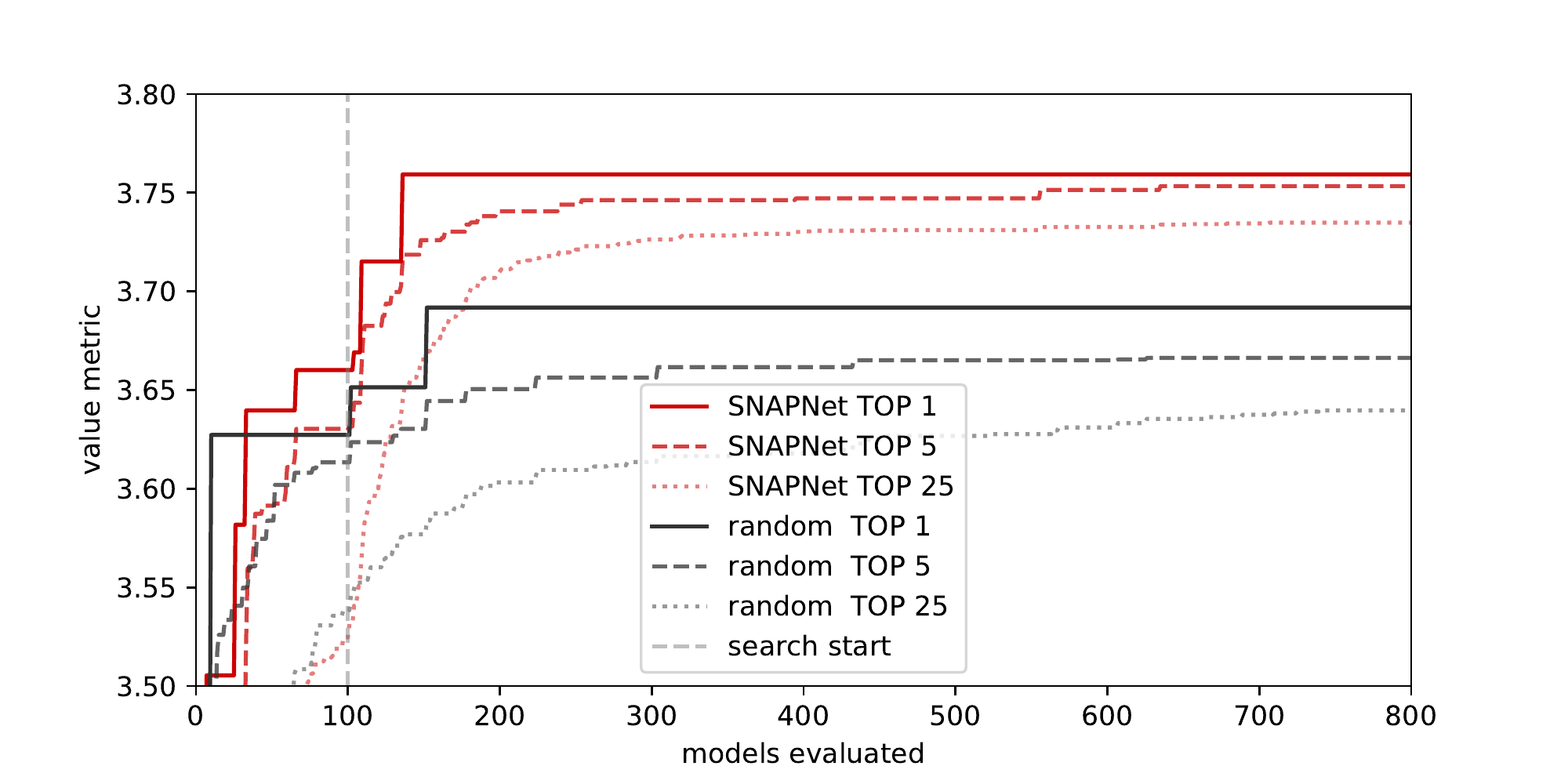}
	}
	~
	\begin{minipage}{47mm}%
		\centering%
		\includegraphics[trim=0 0 0 0, clip,width=47mm]{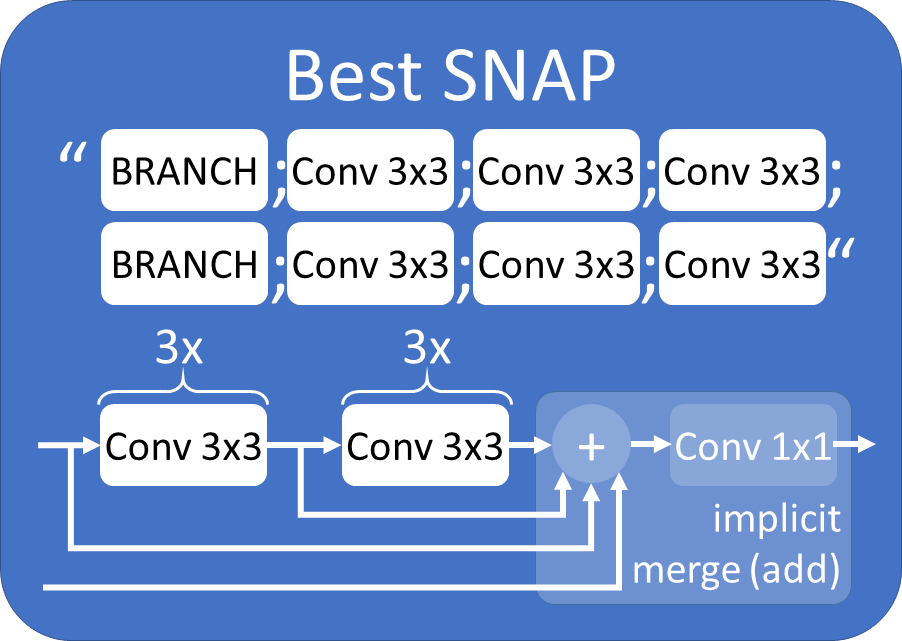}	
	\end{minipage}
	
	\parbox{68mm}{%
		\caption{\label{fig:searchefficiency}Comparison of search efficiency for \mbox{AutoSNAP} (red) and Random Search (black) for the value metric. No significant events occur after 2 GPU days (800 models).}
	}
	~
	\begin{minipage}{47mm}%
		\caption{\label{fig:SNAP}Best SNAP of Auto\-SNAP. Implicit merge of skip-connections, outputs of the previous 2 blocks are used as input.}
	\end{minipage}
\end{figure}
We compare our final architecture (SNAPNet) with two state-of-the-art methods: 
1)~the manually designed i3PosNet \cite{Kugler.2020}, and
2)~an automatically identified architecture using the DARTS$^*$ \cite{Liu.2019} NAS approach.

Both AutoSNAP-based architectures outperform both reference methods by a considerable margin approximately doubling the pose estimation performance.
DARTS-based results do not even reach i3PosNet levels and show the potential of AutoSNAP for CAI applications.
For synthetic images, SNAPNet consistently outperforms position and angle estimates of all other methods including a substantial increase in performance when using i3PosNet's iterative scheme.
For difficult real X-ray images, on the other hand, SNAPNet can significantly reduce the instability of the iterative scheme resulting in a significant reduction of 90\% and 95\% confidence intervals.
In general, \emph{performance gains are slightly more pronounced for real images than for synthetic images}.

The AutoSNAP search strategy is extremely effective, discovering this best performing architecture after less than 10 GPU hours.
We illustrate the top-1 SNAP and neural block used for SNAPNet-A/B in \Cref{fig:SNAP}.
\Cref{fig:searchefficiency} compares the convergence of AutoSNAP to random search (a common NAS baseline), which samples architectures randomly from the search space.
While additional well-performing architectures are discovered after this block confirming flexibility of the approach, 
even the 25th-best SNAPNet architecture outperforms the best architecture produced by random search on the validation set. 
%

%
%




\section{Conclusion}
\label{sec:conclusion}
We propose AutoSNAP, a novel approach targeting efficient search for high-quality neural architectures for instrument pose estimation.
While the application of neural architecture search to CAI is already a novelty, our contribution also introduces SNAPs (to represent block architectures) and an auto-encoder-powered optimization scheme (efficient search algorithm).
This optimization operates on a continuous representation of the architecture in latent space.
We show more than 33 \% error reduction compared to two state-of-the-art methods: the hand-engineered i3PosNet and DARTS, a neural architecture search method.

The application of NAS to CAI is generally limited by a scaling of the  search cost with more operations (\Cref{tab:symbols}) and limited to block optimization, e.g. no macro-architecture optimization.
With respect to learning, AutoSNAP requires stable task evaluations either by good reproducibility or by experiment repetition.
Especially, the identification and exploration of unexplored architectures remains a challenge.

Methods like AutoSNAP enable efficient development and improvement of neural architectures.
It promises to help researchers in finding well-performing architectures with little effort.
In this manner, researchers can focus on the integration of deep neural networks into CAI problems.
While originally designed with instrument pose estimation in CAI in mind, in the future, we will expand AutoSNAP to other CAI and medical imaging problems.

\renewcommand{\bibsection}{\section*{References}} 
\bibliographystyle{splncsnat}
\begingroup
  \small 
  \bibliography{pose-estimation}
\endgroup
%
\end{document}